\documentclass[runningheads,a4paper]{llncs}

\usepackage{amssymb}
\usepackage{graphicx}

\usepackage{bm} 
\usepackage{subfigure} 
\usepackage{amsmath} 

\usepackage{tikz}
\usetikzlibrary{positioning}
\usepackage{booktabs}
 
\usepackage{lipsum}       

\usepackage{footnote}
\makesavenoteenv{tabular}
\makesavenoteenv{table}

\usepackage{hyperref} 

\usepackage{url}
\newcommand{\keywords}[1]{\par\addvspace\baselineskip
\noindent\keywordname\enspace\ignorespaces#1}

\usepackage{floatrow}
\newfloatcommand{capbtabbox}{table}[][\FBwidth]

\begin{document}


\title{Cascaded LSTMs based Deep Reinforcement Learning for Goal-driven Dialogue}

\titlerunning{Cascaded LSTMs based DRL for Goal-driven Dialogue}

%
%
\author{Yue Ma\inst{1} \and Xiaojie Wang\inst{1} \and Zhenjiang Dong\inst{2} \and Hong Chen\inst{2}}

\authorrunning{Y. Ma et al.}

\institute{
Beijing University of Posts and Telecommunications, Beijing, China \\ \email{\{myue, xjwang\}@bupt.edu.cn} \and ZTE Corporation, Nanjing, China \\ \email{dongzhenjiangvip@163.com, chen.hong3@zte.com.cn}
}

\toctitle{Cascaded LSTMs based DRL for Goal-driven Dialogue}
\tocauthor{Y. Ma et al.}
\maketitle

\begin{abstract}
This paper proposes a deep neural network model for jointly modeling Natural Language Understanding and Dialogue Management in goal-driven dialogue systems. There are three parts in this model. A Long Short-Term Memory (LSTM) at the bottom of the network encodes utterances in each dialogue turn into a turn embedding. Dialogue embeddings are learned by a LSTM at the middle of the network, and updated by the feeding of all turn embeddings. The top part is a forward Deep Neural Network which converts dialogue embeddings into the Q-values of different dialogue actions. The cascaded LSTMs based reinforcement learning network is jointly optimized by making use of the rewards received at each dialogue turn as the only supervision information. There is no explicit NLU and dialogue states in the network. Experimental results show that our model outperforms both traditional Markov Decision Process (MDP) model and single LSTM with Deep Q-Network on meeting room booking tasks. Visualization of dialogue embeddings illustrates that the model can learn the representation of dialogue states.

\end{abstract}

\section{Introduction}
A goal-driven dialogue system usually has three components\cite{jurafsky2000speech}: Natural Language Understanding (NLU), Dialogue Management (DM), Natural Language Generation (NLG). Each component includes several subtasks. For example, DM has dialogue state tracking (ST) and action selection (AS). The subtasks are traditionally modeled independently and concatenated in a pipeline way.

There are some important limitations exist in the traditional pipeline system \cite{zhao2016towards}. First, the information cannot be shared between different subtasks due to the separated training methods, for example, the error in the DM could not pass to the NLU. Second, different modules are trained in different ways because they usually use different models, when one of them, for example, NLU is updated with more data, the other modules will fail to adapt to the new parameters in NLU. Third, it is necessary to explicitly define the dialogue states and actions. As the number of slots increases in dialogue tasks, state and action spaces will increase exponentially, this requires significant human efforts.

In order to break these limitations, many researchers have proposed models to deal with the subtasks mentioned above jointly. Some of them jointly modeled subtasks in NLU \cite{guo2014joint,lee2015simultaneous}, some of them jointly modeled subtasks from NLU to ST \cite{henderson2014word}.
Although there are few successful cases on jointly modeling NLU, ST and AS, similar ideas are already applied in computer games. 
Mnih et al. \cite{mnih2013playing,mnih2015human} proposed a deep reinforcement learning (DRL) model for implementing a video game playing agent. By utilizing Deep Q-Network (DQN), screen understanding and game operation selection are blended into an end-to-end model.

Understanding the screen images and text descriptions is similar to NLU, game action selection is similar to AS. The goal of game agent is to achieve maximum long-term rewards \cite{sutton1998introduction} during gameplay, this principle is similarly analogy to the goal in the goal-driven dialogue. Although there are several similarities between a game and a dialogue, a DRL model for game control cannot be simply applied to dialogue control. Different from a game controller, a dialogue agent for goal-driven tasks should not only learn dialogue policies but also track and update a series of explainable dialogue states by merging current utterances with dialogue history.

There are few works on jointly modeling from text input to action selection for goal-driven dialogue in a DRL framework. Narasimhan et al. (2015) \cite{narasimhan2015language} proposed a LSTM based DQN (LSTM-DQN) model for playing text-based games. There are two differences between their model and ours. First, the text provided to the player in the game is the description of its current state. Differently, in goal-driven dialogue, the agent does not have complete information about the environment and the states in dialogues should have been tracked and updated by the agent itself. a single LSTM structure cannot be applied to dialogue domain easily, but the cascaded LSTMs structure in our model can deal with the variable-length dialogues more flexibly and efficiently. 
Zhao and Eskenazi (2016) \cite{zhao2016towards} proposed a framework which is similar to above LSTM-DQN for jointly modeling both state tracking and dialogue policy. The LSTM in their model received a current turn embedding and a history vector as the inputs at each time, while our model employs a shared LSTM to encode both the user utterances and the agent utterances into a turn embedding at each turn, and then use another LSTM to encode all turn embeddings into a dialogue embedding.

Inspired by the proposed ideas about DRL 
\cite{mnih2013playing,mnih2015human,narasimhan2015language}, this paper proposes a cascaded LSTMs based deep reinforcement learning network for building a dialogue agent which jointly models all subtasks through NLU to AS. To the best of our knowledge, it is the first cascaded LSTMs reinforcement learning model. The major contributions of the work are: 1) A cascaded LSTMs structure is designed for firstly encoding user and agent utterances at each dialogue turn into a turn embedding, and then merging turn embeddings into dialogue embedding. Although we do not define any internal delexicalized dialogue states, experimental results showed that dialogue embeddings correlate with some explainable dialogue states well. 2) A deep neural network (DNN) maps dialogue embeddings to the Q-values of different actions. The parameters of cascaded LSTMs and DNN are learned jointly using only the rewards of dialogues. Experimental results showed that our model outperforms a MDP model with fully correct NLU and state tracking and it also outperforms some previous models.

The rest of this paper is organized as follows. The proposed network and its learning algorithm are described in Section 2. Section 3 gives experimental results and analysis. Some conclusions are drawn in Section 4.

\section{Model}
This section is organized as follows. Section 2.1 gives an overview of our model. 
Details of turn embedding and dialogue embedding are described in Section 2.2. 
The learning method is presented at Section 2.3.

\subsection{Overview}

By using all historical information and observations at each moment, our model uses the cascaded LSTMs to model the internal dialogue states without using probabilistic method. The main concepts in our framework are detailed as follows:

1. Let $D=\{D^{1},D^{2},...,D^{T}\}$ be the set of user input utterances, where $D^{t}=d^{t}_{1}d^{t}_{2}...d^{t}_{n}$ is user input at time $t$ (or $t$-th turn), $d^{t}_{i}$ is the $i$-th word of $D^t$.

2. Let $M=\{M^{1},M^{2},...,M^{T}\}$ be the set of agent outputs, where $M^{t}=m^{t}_{1}m^{t}_{2}...m^{t}_{k}$ is the agent output at time $t$ (or $t$-th turn), $m^{t}_{i}$ is the $i$-th word of $M^t$.

3. Let $S$ be the set of states, $s_t$ be the dialogue state at time $t$. 

4. Let $A$ be the set of agent's actions. $a_t$ is the action at time $t$.

5. Let $R$ be the set of rewards, $r_t$ is the reward at time $t$. A small negative reward is assigned for each turn. The last turn of a successful dialogue receives a positive reward.

The goal of the dialogue agent is to find an action $a_{t}$ belonging to $A$ according to state $s_{t}$ at time $t$, which maximizes the expected long-term discounted reward of a dialogue, i.e., to find an optimal policy $\pi:s_{t} \to a_{t}$ satisfying Formula (1).
\begin{align}
Q^*(s,a) = \max\limits_{\pi} E \{ \sum\limits^{\infty}_{t=0} \gamma^{t}r_{t}|s_{t}=s, a_{t}=a, \pi \} \tag{1}
\end{align}
where $\gamma \in (0,1)$ is the discount factor that controls how much the agent prioritizes long-term or short-term rewards, $Q^*$ represents the maximum sum of rewards $r_{t}$ discounted by factor $\gamma$ at each time step. Dialogue state $s_{t}$ should be estimated and tracked before action selection.
\begin{figure}
    \includegraphics[height=1.8in]{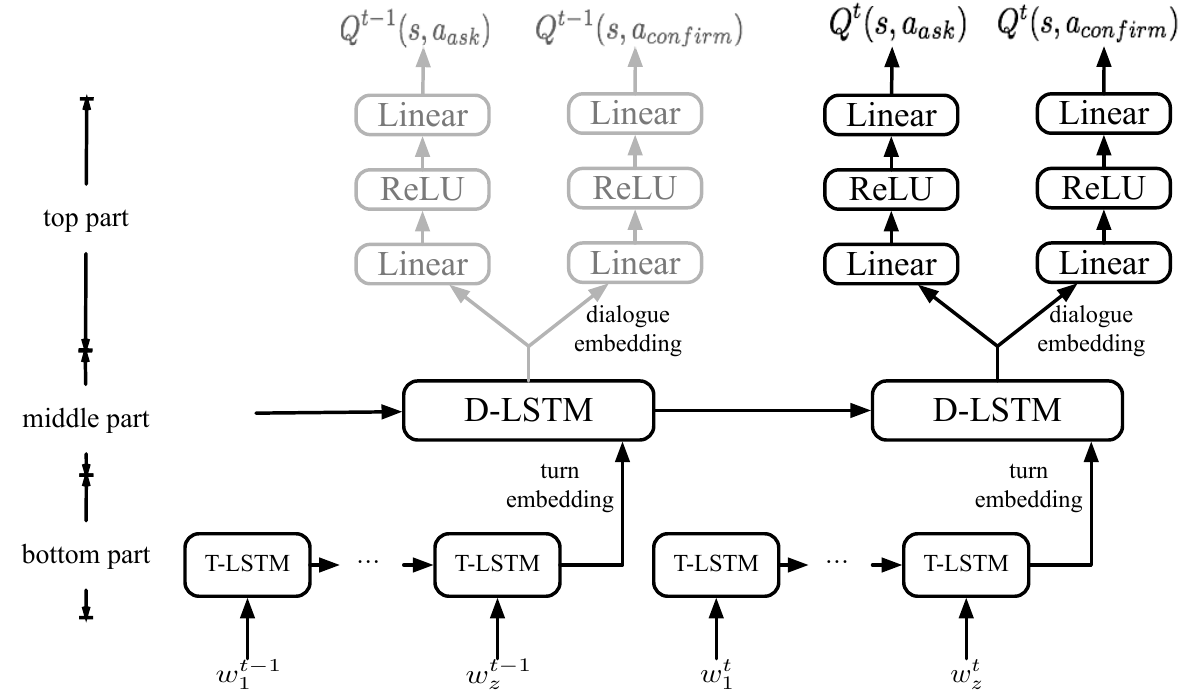}
    \caption[Model]
    {The model structure}
    \label{fig:model}
\end{figure}

The model we propose here jointly models NLU, ST and AS three subtasks. As shown in Fig.~\ref{fig:model}, there are three parts in the network. At the bottom, it is a turn embedding generator which encodes sentences at each turn into a turn embedding. The middle part is dialogue embedding generator which receives previous turn embeddings up to current time and encodes them into a current dialogue embedding. The top part is an action scorer based on DNN which maps dialogue embeddings to the Q-value of each action. All these three parts are cascaded and trained jointly.

\subsection{Turn embedding and Dialogue embedding}

We utilize LSTM \cite{hochreiter1997long} networks as generators of turn embeddings and dialogue embeddings. LSTM is a recurrent network with the ability to connect and recognize long-range patterns between words in text.

At each turn, the same LSTM (named T-LSTM) is used to encode both agent and user utterances into a turn embedding. The input of T-LSTM is a concatenation of agent utterances and user utterances. For $t$-th turn, 
let $I^t = w^t_1w^t_2...w^t_{k}w^t_{k+1}w^t_{k+2} \cdots w^t_{z-1}w^t_{z} = m^t_1m^t_2 \cdots m^t_k\#d^t_1d^t_2 \cdots d^t_n@$
denote the input of T-LSTM, where a symbol ``\#'' is appended to agent word sequence 
and a ``@'' is appended to user word sequence in order to distinguish user and agent utterances. 
$z = n + k + 2$. $w^t_i$ denotes $m^t_i$ for $1 \le i \le k$, and $w^t_i$ denotes $d^t_{i-k-1}$ for $k+2 \le i \le z-1$. Let $h^t = [h^t_1 h^t_2 \cdots h^t_z]$ denotes the hidden states of T-LSTM, $c^t = [c^t_1 c^t_2 \cdots c^t_z]$ is the output of T-LSTM cells. Formulas 2 give details of the network computing.
\begin{align}
    h^t_z, c^t_z = T-LSTM(h^t_{z-1}, w_z) \tag{2}
\end{align}
The final output $c^t_z$ is turn embedding of $I^t$ and then passed to dialogue embedding generator.

A LSTM (named D-LSTM) is used to track dialogue states and generate dialogue embeddings. When a dialogue goes to $T$-th turn, D-LSTM takes $T$ turn embeddings $[c^1_z c^2_z \cdots c^T_z]$ as input and encodes them into a dialogue embedding of $T$ turns. For simplicity, we use $[c_1 c_2 \cdots c_T]$ to denote $[c^1_z c^2_z \cdots c^T_z]$. The output of D-LSTM is $[h'_1 h'_2 \cdots h'_T]$ and its cell output is $[c'_1 c'_2 \cdots c'_T]$. Formulas 3 calculate the output of D-LSTM.
\begin{align}
    h'_T, c'_T = D-LSTM(h'_{T-1}, c_T) \tag{3}
\end{align}

Finally, the top part includes several DNNs. All DNNs share the same dialogue embeddings as network inputs and share the same DNN structure. The number of the DNNs is determined by how many actions it has in dialogue task. For example, if there are actions ``request'' and ``confirm'' in the dialogue, two DNNs should be used in the top part. One DNN stands for one action. The number of output nodes of each DNN is the slot number of the action plus one (used for no act). The outputs of DNN are Q-values of the actions on each slot. Supposing action ``request'' can be followed by 5 slots, then the number of output nodes for its DNN is 6. If the first node is for time slot, then the output of this node is the Q-value of action ``request time slot''. Each DNN selects a slot (or no act) with maximum Q-value separately at each turn. For example, if ``request'' DNN selects time slot and ``confirm'' DNN selects location slot, then a final action ``ask time slot and confirm location slot'' is selected.

By using these three cascaded parts from bottom T-LSTM to top DNNs, the network maps text inputs to dialogue actions end-to-end. At the meantime, dialogue states are kept and updated as dialogue embedding by D-LSTM.

\subsection{Learning method}

Let $\Theta$ be parameters of the model. Double Q-learning \cite{van2016deep} is applied to the proposed network to alleviate the problem of over-estimation. The model keeps two separate networks: a behavior network $\theta$ and a target network $\theta^{-}$. These two networks have the same network structure. After every $L$ updates, the new parameters of $\theta$ are copied over to $\theta^{-}$. At the $j$-th update process, the difference between predicted Q-value $Q(s_{t},a_{t};\theta_{j})$ from the behavior network and the target value $y$ from the target network is used as the error for back-propagation. The target value $y$ is the sum of the reward of taking action $a_t$ at state $s_t$ and the expected Q-value at $s_{t+1}$ which is calculated by $\max\limits_a Q(s_{t+1}, a; \theta^-)$. The detail is shown in Formulas (4.1) and (4.2). Formula (5) gives the gradient of the error.
\begin{align}
    & \, \mathcal{L}(\theta_j) = E[(y - Q(s_t, a_t; \theta_j))^2] \tag{4.1}\\
    & \, y = r_t + \gamma\max\limits_{a_{t+1}}Q(s_{t+1}, a_{t+1}; \theta^-) \tag{4.2}\\
    \frac{\partial \mathcal{L}(\theta_j)}{\partial \theta} = E[2& \, (r_t + \gamma\max\limits_{a_{t+1}}Q(s_{t+1}, a_{t+1}; \theta^-) - Q(s_t, a_t; \theta_j))\frac{Q(s_t, a_t; \theta_j)}{\theta}] \tag{5}
\end{align}

A replay memory pool $X$ is maintained during the training process. Mini-batch transitions are sampled from $X$ at each iteration. Rank-based sampling \cite{schaul2015prioritized} method is used as our selection strategy of transition. The probability of sampling transition $X$ is defined as $P(x) = \frac{p^\alpha_x}{\sum_l p^\alpha_l}$, where $p_x = \frac{1}{rank(x)}$ is the priority of transition $x$. $rank(x)$ is the ranking of transition $x$ in replay memory $X$ which is determined by the deviations. The exponent $\alpha$ determines how much prioritization is used, with $\alpha = 0$ back to the uniform case. At the beginning of the training, the agent interacts with a user simulator by using random action to generate some transitions, and these transitions will be used to initialize the replay memory. After each mini-batch updated, the model will use new transitions to update replay memory following the rank-based strategy.

\section{Experiments}
\subsection{Experimental Settings}

The model is applied to build several dialogue agents. A meeting room booking task is used to 
test the proposed model. The agent communicates with users and gathers required information 
to book a meeting room for users. To book a meeting room successfully, the agent should 
gather values for five required slots in the real-world online system. In order to evaluate the model’s performance comprehensively, four tasks with different number of slots are used. 
They are listed in Table~\ref{table:slotsdetail}. As the number of slots increases, the task becomes more difficult.

\begin{table}
\centering
\caption{Four tasks with different slots}
\begin{tabular}{l|l}
  Slot number & Name of slots\\
  \hline
  2 & start time, location \\
  3 & start time, location, lasting time \\
  4 & start time, location, lasting time \\
  & number of participants \\
  5 & start time, location, lasting time \\
  &number of participants, budget
\end{tabular}
\label{table:slotsdetail}
\end{table}

The inputs of the model are raw utterances come from the agents and user. To train the DRL algorithm, a real-time environment is necessary. The agents built for gameplay are trained by using a game simulator. For dialogue task, it is infeasible to train the agent by communicating with a human in real-time. A user simulator should be used for simulating human behavior and generating user utterances. 

According to the data collected in the online system and the 
basic mode of construction proposed in Li et al. (2016) \cite{li2016user}, several simulators have been built to evaluate the model. Some of the details of the user simulators are as follows:
1) For each slot, there are 25 different slot values and 25 alias values. The alias values are used for simulating errors in Named Entity Recognition (NER), which is named NER-Error. For example, the meeting place which passed to the agent at the first time can be Research Building 09 (the correct one is Research Building 809) (Supplementary A gives an example dialogue).
2) Before giving the right slot value, user simulator first gives an alias value according to the rate of NER-Error. When agents confirm an alias value, the simulator has two choices, one is to reject the confirm action by using general reject templates with a probability of 0.75, and then it waits agents to ask again, the other one is to answer the agents with the true slot value directly.
3) Specified templates are designed for answering questions and for confirming slot values for each slot. Moreover, all slots also share a set of general confirm templates and a set of general reject templates. The simulator can use specified templates for replying agent's questions with a probability of 0.4, or use a general pattern with a probability of 0.6. In order to make the utterances generated by the simulators more complex and make the simulators behave like humans, we collected utterances from the DSTC5 Corpus\footnote{\scriptsize {\tt http://workshop.colips.org/dstc5/data.html}} as a supplement. Utterances which are tagged with {\em RES(POSITIVE)} and {\em FOL(ACK)} in corpus (453 expressions) are used to extend confirm templates. Utterances which are tagged with {\em RES(NEGATIVE)} in corpus (37 expressions) are used to extend reject templates.

The simulators do not set the order of slot filling. It brings more combinations of actions to dialogue, and adds extra complexity to our model to get the optimal policy. In Bordes and Weston (2017) \cite{BordesW16}, the variety of dialogue states is reduced because they have set the order of slot filling.

The agents collect information by asking and confirming slot values. So, two DNNs are employed for action ``ask'' and ``confirm'' respectively. We considered 36 possible command combinations of 6 ask actions and 6 confirm actions. The hyper-parameters of the neural network model are as follows: the word embedding size is 64; the size of turn embedding and dialogue embedding is 64; the size of LSTMs is 128; each DNN has two hidden layers, with 128 nodes in the first layer and the 64 nodes in second layer. Adam optimizer is used for training. The behavior network was updated every 4 steps and the interval between each target network update is 1000 steps. the discount factor $\gamma$ is 0.98. The mini-batch size is 32 and the learning rate is 0.00008.

At each turn, the agent receives a reward of -0.01 if its action is logically acceptable, -1 if not. We do not set a priority level for those acceptable actions. The reward for a successful dialogue is +1 and for an unfinished dialogue is -1. A dialogue ends successfully if an agent get all required slot values from the user simulator within a given number of turns (10 turns, each turn includes no more than 40 words), otherwise, it is unfinished or failed.

We implement the model with Tensorflow. The codes of experiments are available in https://github.com/Damcy/cascadeLSTMDRL. As described in Section 2.3, a replay memory pool is used. By comparing different replay memory sizes, we found a correlation between the agent’s performance and the replay memory size. The replay memory size which is too large or too small is unfavorable to the model performance. We finally employed a replay memory size of 20000 in all experiments.

We first show some comparisons between the model and some previous models, then give some analysis on the model.

\subsection{Model Comparison}

We compared our model with a traditional pipeline agent and LSTM-DQN \cite{narasimhan2015language} on 2, 3 slot tasks as described in above section. The pipeline agent is built with perfect NLU and ST modules, its dialogue state in each turn is fully observable. A tabular Q-learning for MDP is used for action selection. The same $\epsilon-greedy$ strategy is used for MDP method.

For building pipeline agents for tasks with different slot numbers, different sets of states and actions should be explicitly predefined separately. 9 states and 9 actions for 2 slots, 27 states and 16 actions for 3 slots are defined for building 2 slots and 3 slots tasks respectively. For bigger number of slots, the sets of states and actions will increase exponentially. More human efforts will be needed. As for our model, the only necessary change is the number of nodes in DNN's output layer when the model is applied to those tasks with different number of slots.

\begin{table}
\centering
\caption{Results of task with 2 slots and 3 slots}
\begin{tabular}{|c|c|c|c|c|}
    \hline
    & \multicolumn{2}{c|}{2 slots} & \multicolumn{2}{c|}{3 slots} \\ \hline
    & Avg. reward & Avg. length & Avg. reward & Avg. length \\ \hline
    MDP & 0.902 & 4.735 & 0.826 & 5.984 \\
    LSTM-DQN & 0.960 & 4.142 & 0.557 & 6.437 \\
    our model & \textbf{0.961} & \textbf{4.138} & \textbf{0.882} & \textbf{5.622}  \\
    our model* & 0.943 & 4.32 & 0.855 & 5.782 \\
    \hline
\end{tabular}
\label{table:23slots}
\end{table}

Because LSTM cannot deal with long sentences (concatenated each turn utterance in the dialogue, the max input length would be 400), the input of LSTM-DQN could not be the word embeddings. Our model trains the network parameters and word embeddings while LSTM-DQN trains its network parameters and sentence embeddings. Besides the comparison with other models, human evaluation was also carried for our model. Ten graduate students were invited to test the model. Each student completed five dialogues on 2 slots and 3 slots tasks (we showed them all the valid slot values at the beginning). Average length and average total rewards in human evaluation were also reported in evaluation results.
Average dialogue length and average total rewards are calculated on 1000 dialogue steps for three models respectively. And for human evaluation, the averages are obtained from the total number of dialogues.

Table~\ref{table:23slots} show experimental results on 2 and 3 slots tasks respectively, where ``Avg. reward'' is the average of total reward and ``Avg. length'' is the average number of turns for each dialogue. ``our model*'' is the result by human evaluation. It can be seen that our model achieved consistent higher average total rewards and lower average lengths than those in LSTM-DQN agent and MDP method in all tasks. In 2 slot task, LSTM-DQN and our model have similar performance. In 3 slots task, our model outperforms LSTM-DQN significantly. Average total reward of LSTM-DQN falls by 40.3\% from 2 slots to 3 slots, while our model falls by 7.9\%, this shows the robustness of our model. In human evaluation, our model also achieves good performance. Compared with dialogues with simulators, the average total rewards of dialogues with human descends less than 3 percent in both 2 and 3 slots tasks.

\subsection{Hyper-parameter Analysis}

Our model is data-driven. Besides the advantage of reducing human efforts, it is with capabilities of error-toleration and robustness, which will be investigated in this subsection by checking the influence of some parameters.

We give some experimental results on the influence of two parameters. They are the number of slots and NER-Error. Several statistics are used for evaluating the influence. One is the Success Rate of Dialogue (SRD). It is defined in Formula (6).
\begin{align}
    SRD = \frac{\#successful\ dialogues}{\#total\ dialogues\ in\ test} \tag{6}
\end{align}
where a successful dialogue represents that the agent gets all necessary slot values within a given number of turns (The number is set as 10 in all experiments).

The others are No Error Rate (NoER), One Error Rate (OER) and Two Error Rate (TER). NoER is defined in Formula (7). OER is the successful dialogue rate with one action selection error (ASE), TER is the successful dialogue rate with two ASE.
\begin{align}
    NoER &= \frac{\#successful\ dialogues\ without\ ASE}{\#successful\ dialogues} \tag{7}
\end{align}
where an ASE means the agent selects a wrong action. If the agent can find correct response in every turn, then the $NoER = 1$. If the agent can finish a dialogue successfully even with one or more ASEs in it ($NoER < 1$), it means the agent can recover from the errors.

\textbf{Number of Slots}: Number of slots is a task-related parameter. As the number of slots increases, the task becomes more complex. Four models are trained for dialogue tasks with 2, 3, 4 and 5 slots. Table~\ref{table:diffstatistics} shows the SRD and other statistics in different tasks. Our model achieves 100\%, 100\%, 89.06\% and 79.73\% SRD on 2, 3, 4 and 5 slots respectively. The model is sensitive to the complexity of tasks.

\begin{table}
\centering
\caption{Statistics on different slot number and NER-error}
\begin{tabular}{|c|c|c|c|c|}
    \hline
    Slot number \& NER-error &   SRD   &   NoER   &   OER   &   TER  \\
    \hline
    2 slots (15\%)           &  100\%  &  95.52\% &  4.48\% &   0\%  \\
    3 slots (15\%)           &  100\%  &  91.35\% &  8.65\% &   0\%  \\
    4 slots (15\%)           & 89.06\% &  81.25\% &  6.25\% & 1.56\% \\
    5 slots (15\%)           & 79.37\% &  71.43\% &  6.35\% & 1.59\% \\
    2 slots (25\%)           &  100\%  &  94.81\% &  5.19\% &   0\%  \\
    2 slots (35\%)           &  100\%  &  91.30\% &  8.70\% &   0\%  \\
    \hline
\end{tabular}
\label{table:diffstatistics}
\end{table}

\begin{figure}
    \includegraphics[height=1.8in]{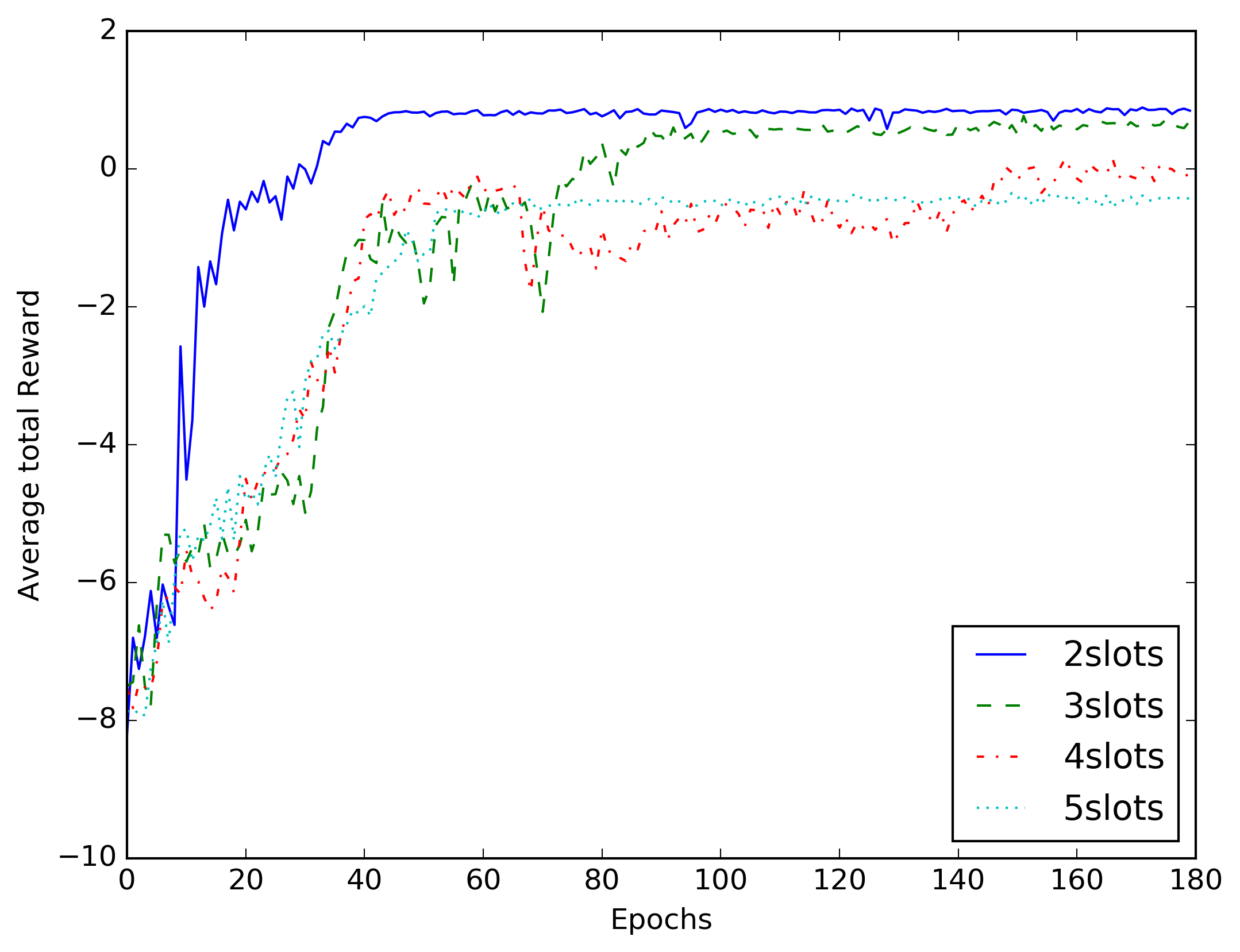}
    \caption[different slot result]
    {Reward curves for different tasks}
    \label{fig:diffslot}
\end{figure}

Results in Table~\ref{table:diffstatistics} also show that our agents can successfully recover from one ASE, and even have chance to recover from two ASEs. For example, in 3 slots task, 91.35\% successful dialogues have no ASE, 8.65\% successful dialogues are recovered from one ASE. In 4 slots task, 6.25\% of successful dialogues include one ASE, and 1.56\% of successful dialogues are recovered from two ASEs. With the increasing of the complexity of tasks, it might be unavoidable for the agent to make mistakes. The more important thing is to recover from mistakes.
Fig.~\ref{fig:diffslot} gives the reward curves for these tasks. More epochs of training will bring bigger rewards in all tasks. Improvements on dialogues with 4 and 5 slots are needed. In another way, we can split a 5 slots task into one 2 slots subtask and one 3 slots subtask according to the independence assumption of slots, the result for SRD was 100\% as well.

\textbf{NER-Error}: Our model not only can recover from its action selection error, but also can deal with errors in NLU. Table~\ref{table:diffstatistics} also shows the experimental results on 2 slots task when different levels of NER-Error are introduced in user simulator. Our model keeps achieving 100\% success rate even when the NER-Error rate is raised from 15\% to 35\%. But it increases the difficulty for leaning the optimal policy. Higher NER-error in training data causes more vibrations and converge slowly (Supplementary B). More NER-error causes higher ASE. The increaseing of OER in Table~\ref{table:diffstatistics} shows our model deals with more ASE with the increaseing of NER-error.

\subsection{Dialogue Embedding Analysis}

\begin{figure}
    \includegraphics[height=1.8in]{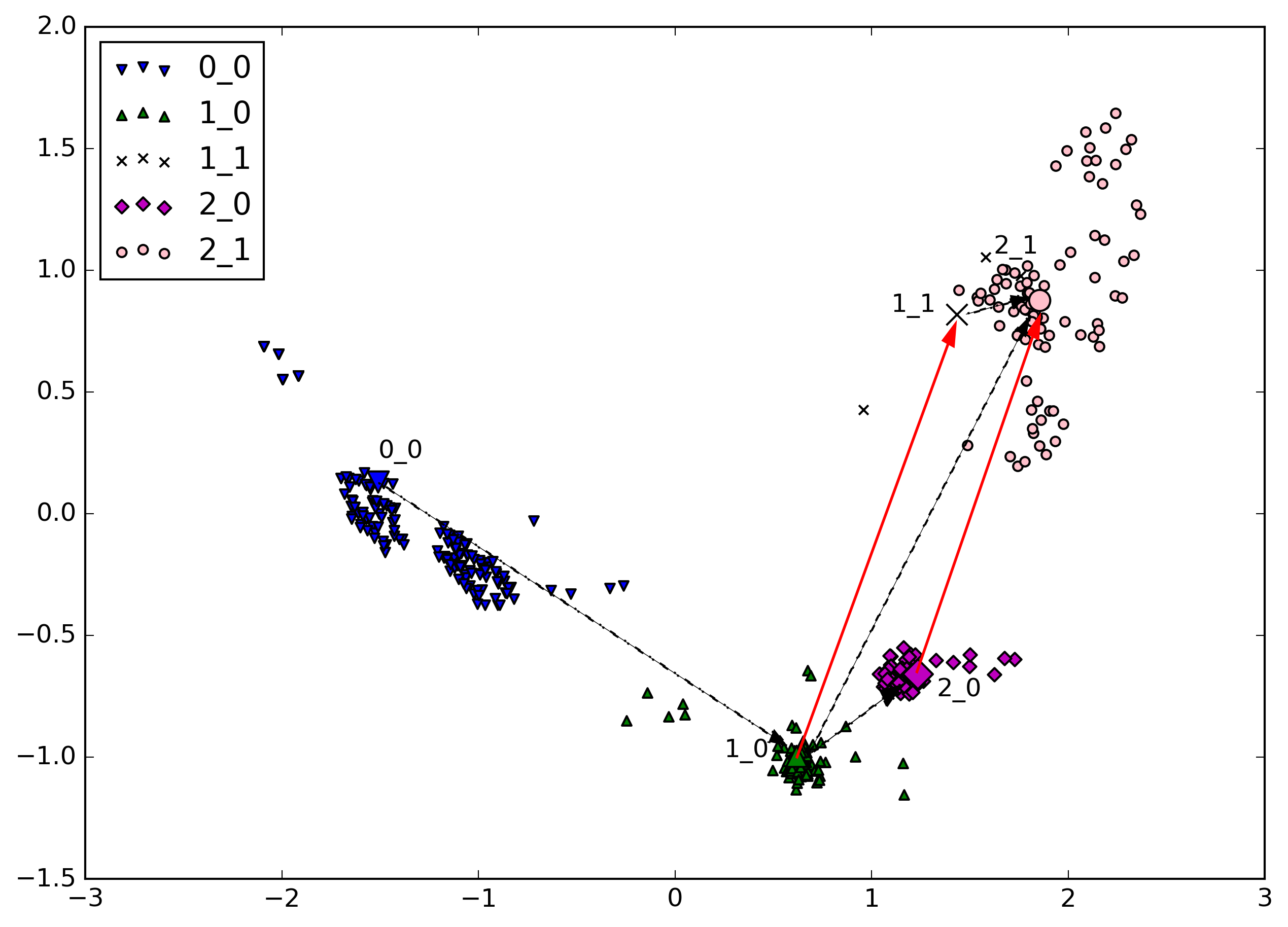}
    \caption[dialogue embedding]
    {Visualization of dialogue embeddings, big symbols are for cluster center points}
    \label{fig:diaemb}
\end{figure}

As mentioned before, the dialogue embedding at $T$-th turn merges all information from previous $T$ turns. PCA projection is used to visualize dialogue embeddings. By using PCA projection, we can see the relation between dialogue embeddings. For 2 slot tasks, let 0\_0 denotes the dialogue state when a dialogue begins with all slot values unknown, 1\_0 denotes the dialogue state when the first slot is given, 2\_0 denotes the dialogue state when the first slot is confirmed, so does the second slot. A subset of dialogue embeddings in test dialogues is collected. The embeddings are then labeled with its state, such as 0\_0, 1\_0 etc. We cluster the embeddings and visualized them in Fig.~\ref{fig:diaemb}. 

From Fig.~\ref{fig:diaemb}, we can find that dialogue embeddings in different dialogues are clustered well. For example, embeddings labeled with 1\_0 in different dialogues are clustered at the bottom of Fig.~\ref{fig:diaemb}, while 2\_1s are clustered at the top right. The number of 1\_1s is small because the agent tends to combine ask action and confirm action at one turn in most of dialogues. This method is better than the one that just confirm the slot what it just asked (it causes the dialogue state transfer from 1\_0 to 2\_0) and then ask a new slot (the dialogue state transfers from 2\_0 to 2\_1), because the latter method needs two turns to achieve the same goal. The agent learns the better strategy by itself through interaction with the user simulator. The agent also find two equivalent ways from 1\_0 to 2\_1. One is from 1\_0 to 1\_1 and then to 2\_1, another way is from 1\_0 to 2\_0 and then to 2\_1. Being well clustered of different dialogue embeddings shows the ability of the model that it can automatically organize context information in different dialogues and implicitly learn the same dialogue states in different dialogues.

The above properties of dialogue embeddings gives an efficient way to identify and explain the dialogue state represented by it. For a dialogue embedding in a dialogue, by identifying which cluster the dialogue embedding belongs to, we can identify which state current dialogue is at. In this way, we explain dialogue embeddings explicity.

\section{Conclusion}

This paper proposes a deep reinforcement learning model for training end-to-end goal-driven dialogue agents. This model uses cascaded LSTMs and DNN structure to model NLU, ST and AS in a single network. The network maps raw utterances to agent actions directly.
Experimental results on meeting room booking tasks show our model outperforms previous models. Visualization of dialogue embeddings illustrates they keep the information of dialogue states.
For dialogue tasks with more slots, the model cannot converge well. More efficient training methods should be explored in the future.

\subsubsection*{Acknowledgments.} This paper is supported by 111 Project (No. B08004), NSFC (No.61273365), Beijing Advanced Innovation Center for Imaging Technology, Engineering Research Center of Information Networks of MOE, and ZTE.


\end{document}